# Efficient Value of Information Computation

Ross D. Shachter
Engineering-Economic Systems and Operations Research Dept.
Stanford University
Stanford, CA 94305-4023
shachter@stanford.edu

## Abstract

One of the most useful sensitivity analysis techniques of decision analysis is the computation of value of information (or clairvoyance), the difference in value obtained by changing the decisions by which some of the uncertainties are observed. In this paper, some simple but powerful extensions to previous algorithms are introduced which allow an efficient value of information calculation on the rooted cluster tree (or strong junction tree) used to solve the original decision problem.

## 1 Introduction

The analysis of sequential decision making under uncertainty is closely related to the analysis of probabilistic inference. In fact, much of the research into efficient methods for probabilistic inference in expert systems has been motivated by the fundamental normative arguments of decision theory. Previous research has applied those developments by modifying algorithms for efficient probabilistic inference on belief networks to address decision making problems represented by influence diagrams (Jensen and others 1994; Ndilikilikesha 1991; Shachter and Ndilikilikesha 1993; Shachter and Peot 1992; Shenoy 1992).

One of the most useful sensitivity analysis techniques of decision analysis is the computation of value of information (or clairvoyance), the difference in value obtained by changing the decisions by which some of the uncertainties are observed (Raiffa 1968). In this paper, some simple but powerful extensions to previous algorithms are introduced which allow an efficient value of information calculation on the rooted cluster tree (or strong junction tree) used to solve the original decision problem.

Dittmer and Jensen(1997) proposed that multiple value of information calculations could all be performed using the same tree. It is this idea that this paper builds on.

Section 2 presents a brief introduction of influence diagrams and Section 3 reviews the most efficient methods for solving them. Section 4 develops some new results which are in applied in Section 5 to efficiently perform multiple value of information calculations. Finally, Section 6 provides some suggestions for future research.

## 2 Influence Diagrams

Influence diagrams are graphical representations for decision problems under uncertainty. In this section the components and notation of influence diagrams are briefly introduced. The graphical structure of the influence diagram reveals conditional independence and the information available at the time decisions must be taken. This is a cursory introduction and the reader is referred to the relevant literature for more information.

An *influence diagram* is a directed graph network representing a single decision maker's beliefs and preferences about a sequence of decisions to be made under uncertainty (Howard and Matheson 1984). The nodes in the influence diagram represent variables– uncertainties (drawn as ovals), decisions (drawn as rectangles), and the criterion values for making decisions (drawn as diamonds). The parents of uncertainties and values condition their distributions, while the parents of decisions represent those variables that will be observed before the decision must be made. The value represents the expected utility of its parents, and decisions are made to maximize this expected utility. When there are multiple value nodes, the total utility is the sum of the utilities for each value. (The results in this paper could also be applied to products (Shachter and Peot 1992; Tatman and Shachter 1990).)

Consider the influence diagram shown in Figure 1 from Dittmer and Jensen(1997). There are four uncertainties, $A, B, C$, and $E$, three decisions, $D_1, D_2$, and $D_3$, and a single value, $U$. The decisions are ordered in the graph and information available at the time of one decision is remembered for subsequent decisions, the *no forgetting* principle. For example, $D_1$ and $C$ are observed before both $D_2$ and $D_3$, while $D_2, E$, and $A$ are observed before only $D_3$. None of the variables are observed before $D_1$ is chosen. Not all of the observations are really needed or *requisite* for a decision. For example, although five of the variables are observed before $D_3$ is chosen, $A$ is the only requisite observation–once $A$ has been observed, the other variables provide no additional information. Similarly, $C$ is the only requisite observation for $D_2$. The diagram can be analyzed to determine the maximal expected utility. If the utility does not represent dollars, we could convert it to dollars by applying the inverse of the utility function that maps from dollars to utility.

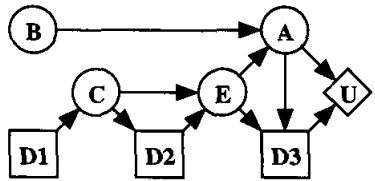

Figure 1: The example influence diagram from Dittmer and Jensen(1997).

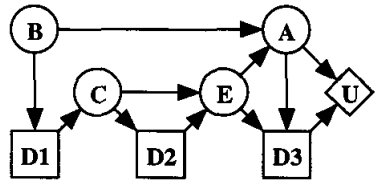

Figure 2: The influence diagram from Figure 1 with clairvoyance on $B$ before $D_1$ is chosen.

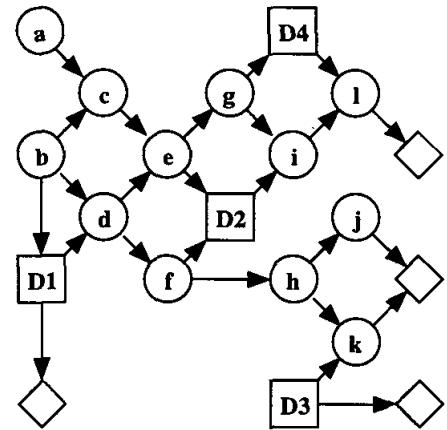

Figure 3: The example influence diagram from Jensen et al(1994).

We can solve a different decision problem without changing any of the distributions in the uncertainties and values by changing the informational assumptions. For example, in Figure 2 $B$ is now observed before $D_1$ is chosen. The expected utility from this diagram must be at least as much as from the earlier diagram because of this extra information, the opportunity to observe $B$. The influence diagram makes it explicit what information is available and when it is available in the two diagrams. This extra value leads to a difference in dollar values called the *value of information* or *value of clairvoyance*. Technically, the value of information is only approximated by this difference (Raiffa 1968), but we will work with this approximated value. Without any new assessments, the decision problem can thus be solved many times, varying the informational assumptions for one variable at a time. This is the process this paper seeks to perform efficiently.

Another influence diagram example that will appear in this paper is shown in Figure 3 (Jensen and others 1994). This diagram has four value nodes, whose functions are summed to obtain the expected utility.

The influence diagram has been developed as a practical representation for a decision problem, and to that end there are several semantic restrictions, which are described in detail elsewhere (Howard and Matheson 1984; Shachter 1986). In particular, we cannot observe the descendant of a decision before making the decision, since the decision can affect its descendants. The one exception is when the descendant represents a constraint and is a deterministic function of the decision and its requisite observations. But this case could be modeled as a value node (with certain cases having prohibitive value) instead of as an observation and thus we can exclude it without loss of generality.

## 3 Rooted Cluster Trees

Efficient algorithms have been developed to solve decision problems represented as influence diagrams. These algorithms build an auxiliary structure called a rooted cluster tree or strong junction tree. Previous work has suggested how value of information calculations could be performed efficiently on such a tree.

Although the influence diagram can be solved directly (Shachter 1986), the most efficient procedures work on related graphical structures (Jensen and others 1994; Ndilikilikesha 1991; Shachter and Ndilikilikesha 1993; Shachter and Peot 1992; Shenoy 1992). This paper considers one of those graphical structures, the rooted cluster tree, a slight generalization of the strong junction tree.

A set of variables is called a *cluster*. A tree of clusters is called a *cluster tree* (or join tree) if every decision or uncertainty appears somewhere in the tree[1], each uncertainty and its parents appear together in at least one cluster, and any variable that appears in two different clusters appears in all of the clusters on the path between them. Corresponding to the notation in Jensen et al(1994), there are two *potential functions* associated with each cluster $C$, a probability potential, $\phi_C$, and a utility potential, $\psi_C$. This paper will introduce and present a minimal amount of this notation, instead focusing on other extensions to Jensen et al(1994). All of the tables in the influence diagram are incorporated into these potential functions.

---

[1] If the cluster tree were not being constructed to compute value of information, it might be worthwhile to exclude variables determined to be extraneous, but here it is desirable to keep all of the variables in the model.

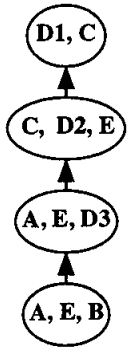

Figure 4: Rooted cluster tree for the influence diagram from Figure 1 from Dittmer and Jensen(1997)

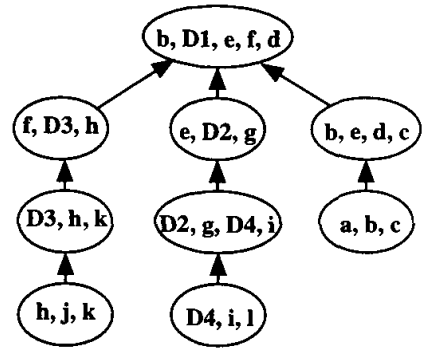

Figure 5: Rooted cluster tree for the influence diagram from Figure 3 from Jensen et al(1994).

The cluster tree is *rooted* if the arcs between clusters are directed so that one cluster, the *root cluster*, has no children, and all of the other clusters have exactly one child. It is useful to distinguish between clusters and variables by their location relative to the root. Cluster $C$ is *inward* of another cluster $C'$ in a rooted cluster tree if $C$ is either the root cluster or between the root cluster and $C'$. In that case $C'$ is said to be *outward* of $C$. If all clusters containing a variable $A$ are outward of some cluster containing a variable $B$ then $A$ is *strictly outward* of $B$ and $B$ is *strictly inward* of $A$. If all clusters containing $A$ either contain $B$ or are outward of a cluster containing $B$, then $A$ is *weakly outward* of $B$ and $B$ is *weakly inward* of $A$. For example in Figure 5, $k$ is strictly outward of $h$, strictly inward of $j$, and neither weakly inward nor weakly outward of $g$.

There are other restrictions that have been developed for rooted cluster trees, but for simplicity only the following, new definition will be presented here. A rooted cluster tree is *properly constructed* for an influence diagram if

1. decision $D$ is strictly inward of decision $D'$ only if $D$ must be chosen before $D'$;

2. decision $D$ is weakly inward of uncertainty $A$ if $A$ is a descendant of $D$ in the influence diagram;

3. decision $D$ is not strictly inward of uncertainty $A$ if $A$ will be observed before $D$ is chosen;

4. decision $D$ and its requisite observations are all contained in some cluster; and

5. any variable $A$ strictly inward of decision $D$ and also in a cluster with $D$ is observed when $D$ is chosen.

Rooted cluster trees properly constructed for the influence diagrams from Section 2 are shown in Figure 4 and Figure 5. The influence diagram's value can then be determined by making a single sweep through the rooted cluster tree toward the root, as summarized in Algorithm 1. The marginalization operator is described in Jensen et al(1994).

**Algorithm 1 (Value Calculation)** *This algorithm computes the optimal expected value on a properly constructed rooted cluster tree.*

*Visit each cluster $C$ in the tree working inward from the leaves toward the root. That is, choose any cluster to visit whose outward neighbors have already been visited. When visiting a cluster, incorporate the updates from $C$'s outward neighbors, and marginalize all variables that do not appear in $C$'s inward neighbor in an order consistent with observation.*

*At the end, the root cluster computes two scalar updates, $\phi_\emptyset$ representing the probability of the evidence and $\psi_\emptyset$, where $\psi_\emptyset/\phi_\emptyset$ is the expected utility of the optimal strategy. For value of information calculations, this latter quantity can be used directly or it can be converted to units of dollars (by applying the inverse utility function).*

This algorithm is generalized in Dittmer and Jensen (1997) to perform multiple value of information calculations with only one cluster tree. The variable(s) to be observed earlier are added to inward clusters. For example, the tree in Figure 6 has uncertainty $B$ added to the three clusters where it did not appear before.

It is not exactly clear how this expanded cluster tree should be processed. According to Dittmer and Jensen (1997), "As mentioned earlier, a control structure is associated with the (strong) junction tree. This structure handles the order of marginalization, and therefore we can use the expanded junction tree (and the associated control structure ) in Figure 7c to marginal-ize $B$ from any clique of our choice. After $B$ has been marginalized from a clique, the table space reserved for $B$ in cliques closer to the strong root is obsolete. Clever use of the control structures will prevent calculations to take place in the remaining table expansions, and the number of table operations in the remaining subtree equals that of an ordinary strong junction tree."

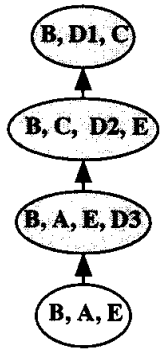

Figure 6: Expanded rooted cluster tree for the influence diagram from Figure 1. Those clusters changed from the rooted cluster tree in Figure 4 are shaded.

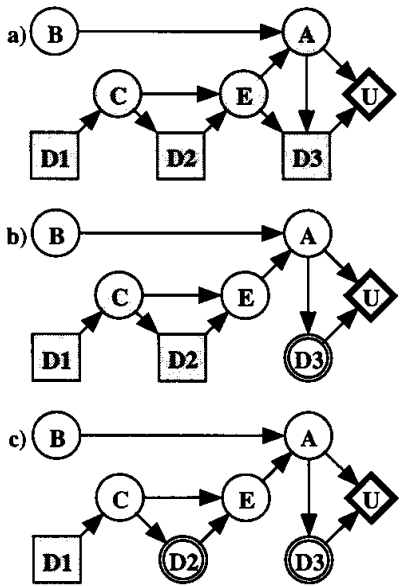

Figure 7: Finding requisite observations for the influence diagram from Figure 1.

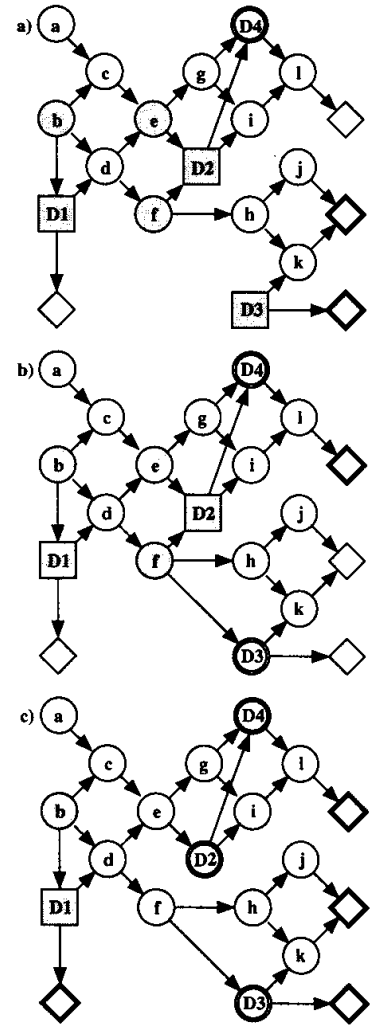

Figure 8: Finding requisite observations for the influence diagram from Figure 3.

## 4 New Results

The definition for a properly constructed rooted cluster tree introduced in Section 3 allows the derivation of some simple but powerful results that will be applied to perform value of information calculations in Section 5. But first it will be helpful to build the best possible rooted cluster trees for the original influence diagram.

The first step in building a cluster tree is recognizing which observations are requisite for the different decisions. Although the BayesBall algorithm (Shachter 1998) is fast (linear time in the size of the graph), it is conservative in computing requisite observations, assuming that the value sets are nested. A less conservative algorithm can be fashioned by teaming BayesBall with the reductions in Tatman and Shachter(1990).

**Algorithm 2 (Requisite Observations)** *This algorithm determines the requisite observations for each decision in an influence diagram as a prelude to proper construction of a rooted cluster tree. It runs in time $O((\text{number of decisions})(\text{graph size}))$.*

*Visit each decision $D_i$ in reverse chronological order, $i = m, \ldots, 1$. Let $V_i$ be the set of value descendants of $D$ in the current diagram. Run the BayesBall algorithm on $V_i$ given $D_i$ and $I_i$, the variables observed before $D_i$ is chosen, and let $R_i$ be the requisite observations (not including $D_i$). Replace $D_i$ by a chance*

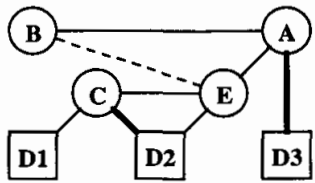

Figure 9: Moral graph based on the modified version of the influence diagram from Figure 1. The value node has been removed, requisite informational arcs are drawn as heavy lines, and the moralizing arc is drawn as dashed line.

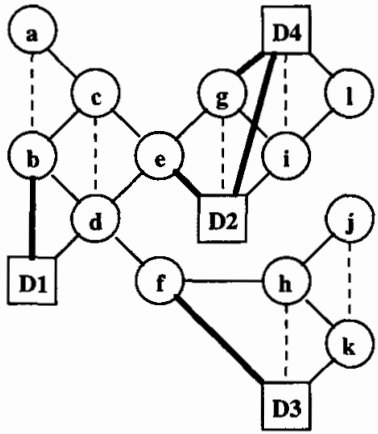

Figure 10: Moral graph based on the modified version of the influence diagram from Figure 3. Value nodes have been removed, requisite informational arcs are drawn as heavy lines, and moralizing arcs are drawn as dashed lines.

node "policy" with $R_i$ as parents and proceed to the next earlier decision.

This algorithm is applied to the two influence diagrams from Section 2, as shown in Figure 7 and Figure 8. In the figures, the value descendants for a particular decision are highlighted, and the decision and its observations are shaded. In Figure 7, it can be seen that the requisite observations are $R_3 = \{A\}$, $R_2 = \{C\}$, and $R_1 = \emptyset$. In Figure 8, the requisite observations are $R_4 = \{g, D_2\}$, $R_3 = \{f\}$, $R_2 = \{e\}$, and $R_1 = \{b\}$. Note the different sets of value descendants.

The next step is to generate the moral graph of the modified diagram. These moral graphs are shown in Figure 9 and Figure 10. The heavy shaded arcs correspond to requisite observations, the dashed lines are moralizing arcs (added between parents with a child in common), and the value nodes have been removed (after any corresponding moralizing arcs were added).

Rooted cluster tree can now be properly constructed based on the moral graphs of the modified diagrams. There is no efficient algorithm to generate such trees, but the structure of the moral graph guides the process (Jensen and others 1994). It can be shown, however, that the method just presented can always yield at least some properly constructed rooted cluster trees.

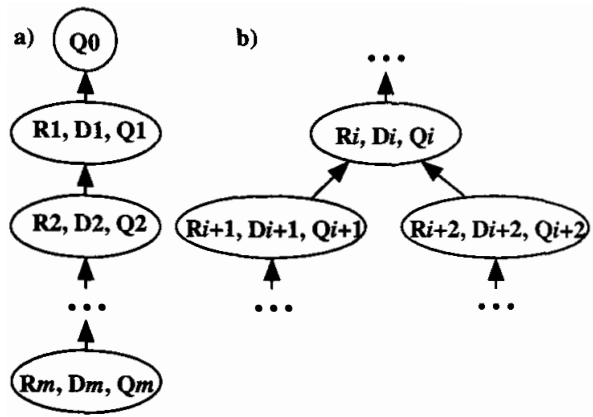

Figure 11: Properly constructed rooted cluster trees.

**Theorem 1 (Requisite Observations)** *Algorithm 2 can be applied to an influence diagram to yield a rooted cluster tree properly constructed for the diagram.*

**Proof:** It is sufficient to show how one such rooted cluster tree could be properly constructed for any influence diagram. At each step of the algorithm, let $Q$ be the non-value variables relevant to $V_i$ as determined by the BayesBall algorithm. (If we are building a rooted cluster tree for potential value of information queries also add to $Q$ any descendants of $D_i$ that could be observed. Otherwise, apparently extraneous variables will not appear in the constructed tree.) Now let $Q_i$ be those nodes in $Q$ for the first time, $Q_i = Q \setminus (Q_{i+1} \cup \ldots \cup Q_m)$. Finally, let $Q_0$ be any nodes relevant to $R_1$ that have not been included in $Q_1, \ldots, Q_m$.

If the value sets are nested, that is, $V_1 \supseteq \ldots \supseteq V_m$, then the rooted cluster tree shown in Figure 11a is properly constructed. Otherwise, $V_{i+1} \not\supseteq V_{i+2}$ if and only if $V_{i+1} \cap V_{i+2} = \emptyset$. Suppose that $V_{i+1} \cap V_{i+2} = \emptyset$ but $V_i \supseteq (V_{i+1} \cup V_{i+2})$. In that case, then the partial tree shown in Figure 11b is properly constructed. □

Of course, the purpose of this exercise is to generate more efficient rooted cluster trees. Examples of such for the influence diagrams from Section 2 are shown in Figure 12 and Figure 13. They are indeed more efficient than the rooted cluster trees in Figure 4 and Figure 5, respectively, reducing the size of the cluster state spaces.

The rest of this section contains the derivation of three simple but powerful results, based on the definition of properly constructed rooted cluster tree. First, the inwardmost cluster with a particular decision must contain its requisite observations.

**Lemma 1 (Current Requisite Observations)** *Given a rooted cluster tree properly constructed for*

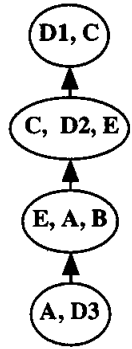

Figure 12: A more efficient rooted cluster tree for the influence diagram from Figure 1.

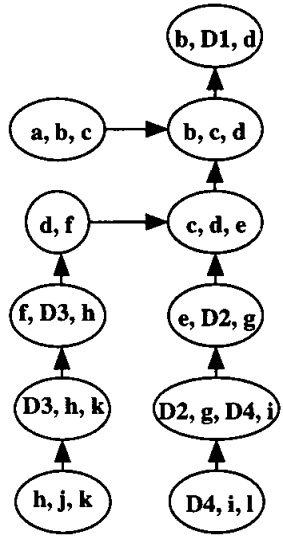

Figure 13: A more efficient rooted cluster tree for the influence diagram from Figure 3.

*an influence diagram, all requisite observed variables for decision D are contained in the inwardmost cluster containing D. Furthermore, any variables in both that cluster and the next inward cluster are observed when D is chosen.*

**Proof:** By proper construction, any variable observed before $D$ is chosen must be weakly inward of $D$ and any requisite observation must be contained in a cluster with $D$. On the other hand, if $A$ is not observed before $D$ is chosen and strictly inward of $D$ then it must not be contained in that cluster. □

Next, when an uncertainty becomes observable before decision $D$ is chosen, it will not become requisite unless it is weakly outward to $D$.

**Theorem 2 (Newly Requisite Observations)** *Given a rooted cluster tree properly constructed for an influence diagram, if uncertainty A is not weakly outward of decision D nor in any clusters with D then if A were to be observed before D were chosen it would*

*not be requisite for D.*

**Proof:** By proper construction and Lemma 1, the utility from $D$ is weakly outward from $D$ and all variables in common between the inwardmost cluster containing $D$ and the next inward cluster are observed when $D$ is chosen. Therefore, the utility is separated in the cluster tree from $A$ by observations for $D$, and the utility from $D$ is conditionally independent of $A$ given the observations for $D$ (Jensen and others 1990a; Lauritzen and others 1990). □

Finally, when an uncertainty stops being observable before decision $D$ is chosen, all of the observations now requisite for $D$ are weakly inward.

**Proposition 1 (Previously Requisite Observations)** *Given a rooted cluster tree properly constructed for an influence diagram where uncertainty A is observed before decision D is chosen, then if A were not to be observed before D were chosen, all variables which would be requisite observations for D are weakly inward of D.*

**Proof:** When properly constructed, all variables observed before decision $D$ is chosen (not just the requisite ones) are weakly inward of decision $D$. □

## 5 Computing the Value of Information

The new results from Section 4 can now be applied to perform value of information calculations on the rooted cluster tree for the original influence diagram. First a method is presented for computing the value of a decision problem when an uncertainty is already observed. This is then generalized to computing the value when there is an earlier observation, and then when there is a later observation.

Suppose that an uncertainty has already been observed, such as $a$ in the influence diagram shown in Figure 3. By Theorem 2 it can be requisite only for decisions weakly inward in the tree shown in Figure 13. By exploiting the probabilistic heritage of the decision algorithm (Jensen and others 1990b; Lauritzen and Spiegelhalter 1988), the rooted cluster tree is ideally suited to solve this problem. First, the evidence is stored in the probability potential in a cluster containing $a$, say the inwardmost cluster containing $a$. Now Algorithm 1 could be run incorporating this evidence.

Suppose, however, that Algorithm 1 had already been run before $a$ was observed. No problem–only the clusters between the inwardmost cluster containing $a$ and the root need to be visited. All of the other calculations are unchanged! We could perform this same operation even if $a$ were not observed precisely, provided we had some imperfect observation about $a$ represented by a likelihood function.

Now consider the case in which an uncertainty will be observed earlier, but has not yet been observed, such as $B$ in Figure 2. Again it is possible to exploit the well-known properties of cluster trees. To compute the value of the decision problem in which $B$ will be observed earlier, cycle through all of the possible values of $B$, performing the calculations each time as though $B$ were observed. The potentials computed can then be summed, thereby incorporating the probability distribution over the different possible values of $B$. If this summing occurs immediately after the optimal policy for $D_i$ is computed, then this is the value of observing $B$ before $D_i$ is chosen.

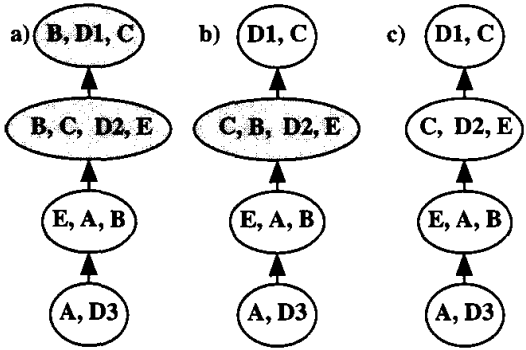

Figure 14: Effective rooted cluster trees for value of information calculations on the influence diagram from Figure 1 when $B$ is observed before decisions are made. Those clusters changed from the rooted cluster tree in Figure 12 are shaded.

One can think of this as "effectively" adding $B$ to the clusters inward to the inwardmost cluster containing $D_i$, as shown in Figure 14. In Figure 14a, this is used to compute the value of observing $B$ before $D_1$, and in Figure 14b before $D_2$. Figure 14c is not different from Figure 12 because $B$ would not be requisite if it were observed before $D_3$. This can be recognized immediately from the rooted cluster tree in Figure 12 in which $B$ is inward of $D_3$. Note that unlike Figure 4, in which the tree has been "expanded," this approach sums over cases, doing the same work, but there is no need to store the larger tables, and it uses the original rooted cluster tree!

Now consider the influence diagram shown in Figure 3. Observing $a$ earlier yields the effective rooted cluster tree shown in Figure 15a. Uncertainty $a$ would be requisite for $D_1$ but not for any of the later decisions. Suppose instead that $j$ were observed earlier. It cannot be observed before $D_1$ since it is a descendant of $D_1$. It is not requisite for $D_2$ or $D_4$ since it is not inward of either, but it would be requisite for $D_3$ as shown in Figure 15b.

Now suppose that a variable is observed later rather than earlier. Consider $C$ in Figure 1 and suppose that it is no longer observed before $D_2$ is chosen. From Proposition 1, the observations now requisite for $D_2$ are inward, so the solution is to run Algorithm 2 to

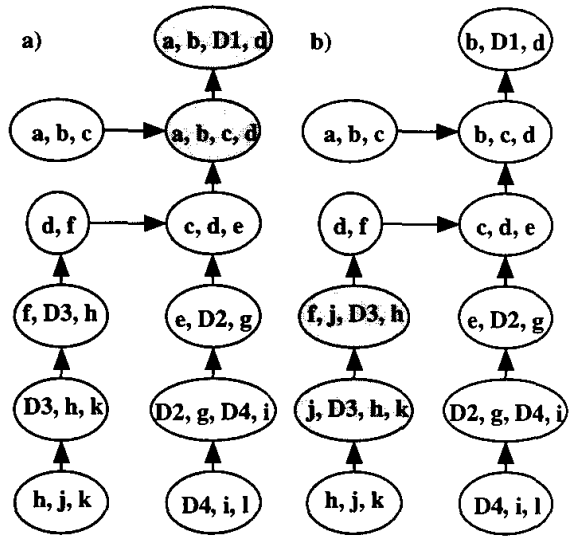

Figure 15: Some effective rooted cluster trees for value of information calculations on the influence diagram from Figure 3.

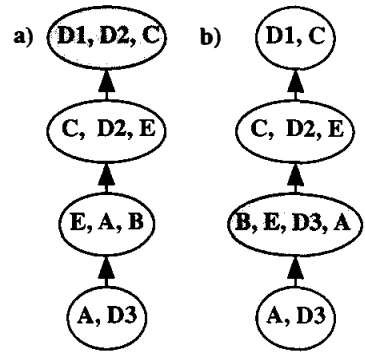

Figure 16: Effective rooted cluster trees for value of information calculations on the influence diagram from Figure 1 when the observation of either $C$ or $E$ is delayed.

figure how inward $D_2$ must effectively move up as in Figure 16a. Only now maximize over different cases for $D_2$ instead of summing. Similarly, if $A$ were not observed for $D_3$, $D_3$ can be effectively moved inward as in Figure 16b. Finally, if $E$ were not observed for $D_3$ there is no change, since $E$ is not requisite for $D_3$.

Finally, a similar process can be done for the diagram in Figure 3. Figure 17a shows the effective rooted cluster tree when $f$ is no longer observed before $D_3$ and Figure 17b shows the effective tree when $e$ is no longer observed before $D_2$.

## 6 Conclusions and Future Research

This paper has developed improved value of information calculations over previous work in two respects. First, it improves the rooted cluster trees used to solve for the value of a decision problem. Second, it develops methods for reusing the original tree in order to perform multiple value of information calculations.

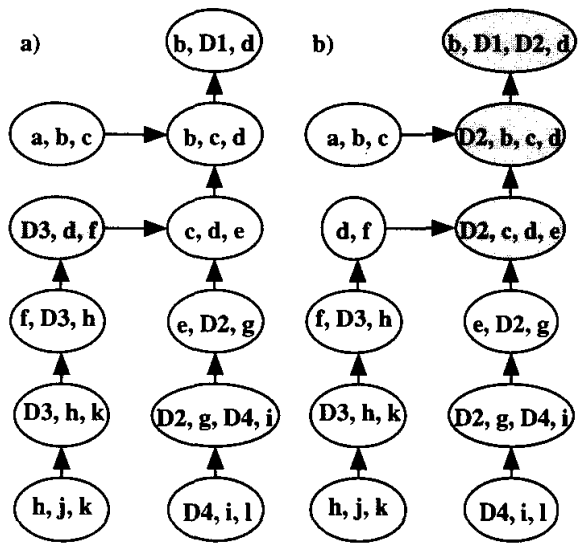

Figure 17: Some effective rooted cluster trees for value of information calculations on the influence diagram from Figure 3.

There are several opportunities for further research. When a particular variable is observed at multiple earlier decisions it should be possible to reuse some of the calculations. Also, this approach exploits the special properties of changing the time when a single uncertainty becomes observed. It would be useful if the method could be generalized to solve the decision problem with any set of informational assumptions from the original rooted cluster tree. If that could be done efficiently, then the original decision problem could be solved from the most convenient cluster tree.

## Acknowledgments

This paper has benefited from the comments, suggestions, and ideas of friends and students, most notably Mark Peot and Prakash Shenoy.